%% file: 0_Main_IEEE.tex
\documentclass[journal]{IEEEtran}

\usepackage[T1]{fontenc}
\usepackage[utf8]{inputenc}   
\usepackage{cite}
\usepackage{amsmath,amssymb,amsfonts}
\usepackage{graphicx}
\usepackage{booktabs}
\usepackage{multirow}
\usepackage{array}
\usepackage{tabularx}
\usepackage{caption}
\usepackage{enumitem}
\usepackage{xspace}
\usepackage{xcolor}

\graphicspath{{image/}{figs/}{./}}

\newcolumntype{Y}{>{\raggedright\arraybackslash}X}
\newcolumntype{P}[1]{>{\raggedright\arraybackslash}p{#1}}
\renewcommand{\arraystretch}{1.1}
\newcommand{\framework}{SRICL\xspace}

\title{Job Skill Extraction via LLM-Centric Multi-Module Framework}

\author{
Guojing~Li$^{1,2*}$, Zichuan~Fu$^{1*}$, Junyi~Li$^{1}$, Faxue~Liu$^{1}$,Wenxia~Zhou$^{2}$,
Yejing~Wang$^{1}$, Jingtong~Gao$^{1}$, Maolin~Wang$^{1}$, Rungen~Liu$^{1}$, Wenlin~Zhang$^{1}$,
and Xiangyu~Zhao$^{1\dagger}$%
\thanks{$^{1}$City University of Hong Kong.}%
\thanks{$^{2}$Renmin University of China.}%
\thanks{$^{*}$Guojing Li and Zichuan Fu contributed equally to this work.}%
\thanks{$^{\dagger}$Corresponding authors: Guojing Li (e-mail: guojingli3-c@my.cityu.edu.hk) and Xiangyu Zhao (e-mail: xianzhao@cityu.edu.hk).}%
}

\begin{document}

\maketitle

\begin{abstract}
Span-level skill extraction from job advertisements underpins candidate--job matching and labor-market analytics, yet generative large language models (LLMs) often yield malformed spans, boundary drift, and hallucinations, especially with long-tail terms and cross-domain shift. We present \framework, an LLM-centric framework that combines semantic retrieval (SR), in-context learning (ICL), and supervised fine-tuning (SFT) with a deterministic verifier. SR pulls in-domain annotated sentences and definitions from ESCO to form format-constrained prompts that stabilize boundaries and handle coordination. SFT aligns output behavior, while the verifier enforces pairing, non-overlap, and BIO legality with minimal retries. On six public span-labeled corpora of job-ad sentences across sectors and languages, \framework achieves substantial STRICT-F1 improvements over GPT-3.5 prompting baselines and sharply reduces invalid tags and hallucinated spans, enabling dependable sentence-level deployment in low-resource, multi-domain settings.
\end{abstract}


\input{1Introduction}
\input{2Framework}

\input{3Experiments}
\input{5RelatedWork}

\input{6Conclusion}
\appendices
\input{7Appendix}

\bibliographystyle{IEEEtran}
\bibliography{9Reference}

\end{document}

%% file: 1Introduction.tex
\section{Introduction}
 The extraction of job skills from job advertisements is a span-level information extraction (IE) task under BIO tagging that supports recruitment matching, recommendation, and labor market analysis. While large language models (LLMs) enable low-overhead IE via prompting, deployment at scale is hampered by long-tail, domain-specific terminology, multi-token coordination, and cross-industry/domain shift \cite{zhang2022skillspan,senger2024deep}. Beyond the usual precision–recall trade-offs, production systems must ensure {BIO validity} and {controllability}, avoiding illegal tags, boundary drift, and hallucinated skills, that are often under-addressed in generic LLM pipelines \cite{manakul2023selfcheckgpt,xu2024large}.

Existing approaches fall into three families. (i) Supervised neural NER (e.g., BiLSTM-CRF and pretrained encoders)~\cite{conneau-etal-2020-xlm-r,he2023unleashing} achieves strong in-domain scores but is label-hungry, brittle under domain/language transfer, and prone to missing compositional or tail skills. (ii) \emph{Prompt-based LLMs}~\cite{ma2023large,manakul2023selfcheckgpt} lower annotation cost yet may hallucinate non-existent skills and blur token-aligned boundaries; without explicit grounding or verification, few-shot gains are unstable across datasets. Even with reasoning strategies such as Chain-of-Thought (CoT) and self-consistency, generic prompting seldom guarantees span-level robustness \cite{wei2022cot,wang2022self}. (iii) \emph{Retrieval-augmented variants}~\cite{lewis2020rag} improve coverage, but one-shot retrieval and generic CoT rarely translate into cleaner span boundaries or guaranteed BIO legality. Unified IE frameworks (e.g., UIE/InstructUIE)~\cite{lu-etal-2022-unified,wang2023instructuie} report promising text-to-structure results, yet reliability at BIO granularity for skill spans remains open.

\begin{figure*}[t]
  \centering
  \includegraphics[width=\textwidth]{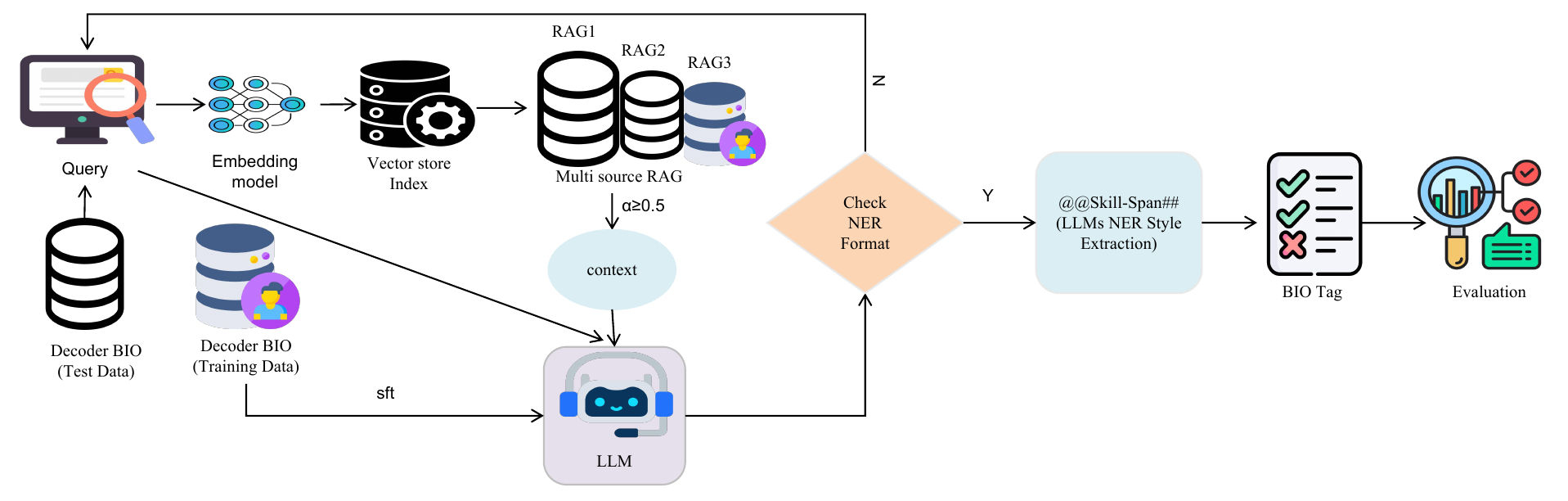}
  \caption{The overall architecture of our proposed SRICL framework.} 
  \label{fig:pipeline_overview}
\end{figure*}

To address these limitations, we propose \textbf{SRICL}, an LLM-centric framework defined as \textbf{SFT + RAG + In-Context Learning}.
\emph{(1) Supervised Fine-Tuning (SFT).} We fine-tune an open-source LLM for span-level tagging under task-specific prompts, which improves boundary sensitivity and reduces reliance on heavy few-shot priming.
\emph{(2) Retrieval-Augmented Generation (RAG).} We adopt a multi-\emph{source} design: \textbf{RAG-1} retrieves \emph{in-domain annotated sentences} as dynamic demonstrations; \textbf{RAG-2} retrieves \emph{authoritative definitions} (e.g., ESCO) for sense disambiguation and boundary regularization; \textbf{RAG-3} retrieves near-neighbor hard cases across sectors/languages to enhance robustness under domain shift. A lightweight gating/weighting mechanism balances these sources per input.
\emph{(3) In-Context Learning (ICL).} We construct a modular ICL prompt that explicitly includes Prompt Engineering (instruction and format binder), a System Persona (a constrained ``skill-span extractor'' with refusal and legality policies), and a precise Task Definition (BIO schema, boundary rules, and output contract). Retrieved evidence from RAG-1/2/3 is assembled into the prompt as demonstrations and definitions. We pair this with format-constrained decoding to stabilize token-aligned boundaries and mitigate coordination errors. For reliability control, we adopt a previously proposed checking-and-retry verifier~\cite{kim2024verifiner,madaan2023self,shinn2023reflexion} and integrate it via deterministic orchestration to ensure tag-scheme validity (round-trip between @@\#\# and BIO), correct boundary inconsistencies, and suppress hallucinated spans.

We evaluate on six public datasets covering multiple sectors and languages.  Our SRICL method consistently improves STRICT-F1 over strong GPT-3.5 prompting baselines while substantially lowering illegal-BIO and hallucination rates. The ablation study shows that RAG-1/2/3 chiefly contribute to recall and domain robustness; ICL (with persona, task definition, and prompt engineering) stabilizes boundaries and reduces coordination errors; the verifier ensures scheme compliance and curbs rare but high-impact hallucinations. 

%% file: 2Framework.tex
\section{Proposed Method}
\label{sec:method}

\subsection{Problem Formulation}
Given a job-ad sentence $x=(w_1,\dots,w_T)$, the goal is to assign span-level BIO tags $y\in\{\text{Beginning},\text{Inside},\text{Outside}\}$. Outputs must be boundary-accurate, BIO-legal, and avoid hallucinated skills not supported by the text or target taxonomy.

\subsection{Overview}
We target \emph{reliability} and \emph{controllability} with four modules (Fig.~\ref{fig:pipeline_overview}):
(i) \textbf{SFT with span anchoring} on an open LLM;
(ii) \textbf{data-specific prompting} with BIO-only decoding;
(iii) \textbf{multi-source retrieval (RAG)} from in-domain sentences and authoritative skill/occupation definitions; and
(iv) \textbf{verification with targeted retry} to enforce BIO legality and curb hallucination.
The design remains plug-in friendly to domain KBs and multi-task extensions.

\subsection{SFT with Span Anchoring}
We convert BIO labels to anchored targets by bracketing each gold span with \verb|@@...##|, producing
$(\texttt{Sentence}:x,~\texttt{Target}:\widetilde{x})$
for instruction-style SFT (LoRA on Qwen2.5-14B). To harden the model, we use
(1) \emph{span-balanced sampling} to reduce head-skill bias;
(2) \emph{connector normalization} (e.g., “A/B”, “X, Y and Z”) for cleaner boundaries; and
(3) \emph{hard negatives} (all-O sentences).
This raises boundary precision and stabilizes format with a compact backbone.

\subsection{Dataset-Specific Prompting and BIO-only Decoding}
Each dataset uses a tailored prompt encoding language, domain jargon, and typical span shape. A fixed \texttt{system} segment defines the task and skill notion (aligned with ESCO); the \texttt{user} segment supplies the raw sentence (or a token list). We constrain decoding to \emph{BIO-only} to keep outputs concise and controllable.

\subsection{Multi-Source Retrieval (RAG)}
For each $x$, we retrieve:
\begin{itemize}[leftmargin=*,itemsep=0pt]
\item \textbf{RAG-1:} in-domain annotated sentences as few-shot demonstrations (phrase and boundary style matching);
\item \textbf{RAG-2:} authoritative definitions (e.g., ESCO) for disambiguation and boundary regularization.
\end{itemize}
Both sources are dense-indexed; candidates are selected by cosine similarity $s(q,d)=\langle f(x),f(d)\rangle$ and inserted as compact snippets. When available, a domain KB (\textbf{RAG-3}) can be plugged in without retraining to cover cold-start verticals and fast-evolving terminology.

\subsection{Verification and Targeted Retry}
A deterministic checker validates (i) BIO legality (no I-after-O, no crossing/nesting) and (ii) semantic/taxonomy consistency. On failure, we re-prompt with targeted hints under a small retry budget $K\in\{1,3\}$. The checker is lightweight, model-agnostic, and complements retrieval.

\begin{table}[t]
  \centering
  \footnotesize
  \setlength{\tabcolsep}{6pt}
  \caption{Corpus snapshot (trimmed to the most informative metric). \#Train, \#Test, \#Train-Skill, and \#Test-Skill denote the number of instances in training and test set, and the number of distinct skills in training and test set.}
  \label{tab:corpus-stats}
  \begin{tabular}{llrrr}
    \toprule
    Dataset & \#Train & \#Test & \#Train-Skill & \#Test-Skill \\
    \midrule
    {SkillSpan}   & 4782 & 3569 & 1805 & 984 \\ 
    {Kompetencer} & 778 & 262 & 365 & 102\\
    {Sayfullina}  & 3705 & 1851 & 803 & 580 \\
    {Gnehm}       & 19824 & 2550 & 4262 & 674 \\
    {Green}       & 8668 & 335 & 8713 & 582 \\
    {FIJO}        & 399 & 50 & 515 & 122 \\
    \bottomrule
  \end{tabular}
\end{table}


%% file: 3Experiments.tex
\section{Experimental Results and Analysis}
\label{sec:typestyle}

\subsection{Experimental Setup}

\subsubsection{Datasets and Metrics}

We evaluate on six public datasets spanning multiple languages and domains: \textit{SkillSpan}, \textit{Kompetencer}, \textit{Green}, \textit{FIJO}, \textit{Sayfullina}, and \textit{GNEHM}.
Corpus-level descriptive statistics (train/test size, average skill density, sentence length, long-tail coverage) are summarized in Table~\ref{tab:corpus-stats}.
Following prior work, we report token-level Precision (P), Recall (R) and F1 under the \textsc{Strict} span matching protocol.

\subsubsection{Comparison Setting}
Our main table reports results under a unified generative BIO-only decoding interface (sentence → generated BIO tags) with the same output contract, optionally augmented by retrieval and verifier/retry. Encoder-based token-classification models that rely on supervised fine-tuning (e.g., JobBERT-style models trained/adapted with large-scale domain data) follow a different training and inference paradigm; we therefore discuss/cite them separately as reference points rather than treating them as direct main-table baselines.

\subsubsection{Baselines}
We compare against (i) GPT-3.5 zero-/few-shot prompting baselines (with/without kNN and data-specific prompts), and (ii) open-source LLaMA-3-8B and Qwen2.5-14B under identical prompting/decoding settings. Unless otherwise noted, the deterministic verifier is disabled to equalize latency; its impact is reported separately.



\subsection{Main Results}
Table~\ref{tab:main_experiment} lists token-level \textsc{P/R/F1} on the six datasets.
Overall, our Qwen2.5-14B implementation attains the best average F1 among open-source LLM variants and consistently outperforms GPT-3.5/zero-/few-shot settings.

\begin{table*}[t]
  \centering
  \caption{Main Experiment (STRICT \& RELAX). Metrics are in \%.}
  \label{tab:main_experiment}
  \setlength{\tabcolsep}{3.5pt}
  \renewcommand{\arraystretch}{1.08}
  \resizebox{\textwidth}{!}{%
  \begin{tabular}{@{}l*{7}{ccc}@{}}
    \toprule
    \textbf{Method}
    & \multicolumn{3}{c}{\textbf{SKILLSPAN}}
    & \multicolumn{3}{c}{\textbf{KOMPETENCER}}
    & \multicolumn{3}{c}{\textbf{GREEN}}
    & \multicolumn{3}{c}{\textbf{FIJO}}
    & \multicolumn{3}{c}{\textbf{SAYFULLINA}}
    & \multicolumn{3}{c}{\textbf{GNEHM}}
    & \multicolumn{3}{c}{\textbf{AVG}} \\
    \cmidrule(lr){2-4}\cmidrule(lr){5-7}\cmidrule(lr){8-10}\cmidrule(lr){11-13}\cmidrule(lr){14-16}\cmidrule(lr){17-19}\cmidrule(lr){20-22}
    & \textbf{P} & \textbf{R} & \textbf{F1}
    & \textbf{P} & \textbf{R} & \textbf{F1}
    & \textbf{P} & \textbf{R} & \textbf{F1}
    & \textbf{P} & \textbf{R} & \textbf{F1}
    & \textbf{P} & \textbf{R} & \textbf{F1}
    & \textbf{P} & \textbf{R} & \textbf{F1}
    & \textbf{P} & \textbf{R} & \textbf{F1} \\
    \midrule
    \multicolumn{22}{l}{\textbf{STRICT / NER-STYLE}} \\
    zero-shot+specific (GPT-3.5)
    & 3.08 & 11.94 & 4.90
    & 1.54 & 4.76 & 2.33
    & 7.89 & 2.10 & 3.32
    & 21.43 & 9.68 & 13.33
    & 5.11 & 3.42 & 4.10
    & 5.92 & 13.80 & 8.29
    & 6.42 & 7.62 & 6.04 \\
    5-shot
    & 35.46 & 29.50 & \textbf{32.21}
    & 27.27 & 25.00 & \textbf{26.09}
    & 24.53 & 8.72 & 12.87
    & 20.00 & 9.09 & 12.50
    & 30.35 & 36.88 & 33.20
    & \textbf{56.16} & 53.72 & \textbf{54.91}
    & 27.68 & 27.15 & 28.63 \\
    +kNN
    & 26.96 & 23.17 & 24.92
    & 10.55 & 23.53 & 22.22
    & 22.86 & 18.31 & 20.88
    & 41.18 & 31.82 & 35.90
    & 24.97 & 30.08 & 27.29
    & 55.31 & 48.48 & 51.67
    & 25.98 & 29.23 & 30.48 \\
    +kNN+specific
    & 10.89 & 24.48 & 15.08
    & 10.55 & 20.79 & 14.00
    & 30.02 & 27.30 & \textbf{28.59}
    & 55.31 & 48.48 & \textbf{51.67}
    & 26.60 & 32.63 & 29.31
    & 51.89 & 52.92 & 52.40
    & 26.47 & 34.43 & 31.84 \\
    GPT-3.5 Version (+kNN+specific)
    & 9.10 & 26.60 & 13.70
    & 11.20 & 29.20 & 16.10
    & 30.80 & 32.80 & 31.80
    & 43.70 & 44.70 & 44.20
    & 29.80 & 47.30 & 36.60
    & 29.90 & 64.80 & 40.90
    & 25.75 & 40.90 & 30.55 \\
    Llama-3-8B-SRICL(B8-Our Method)
    & 31.50 & 55.09 & 40.08
    & 42.11 & 45.71 & 43.84
    & 23.98 & 20.98 & 22.38
    & 23.94 & 27.64 & 25.66
    & 22.15 & 29.03 & 25.13
    & 50.93 & 52.41 & 51.66
    & 32.43 & 38.48 & 34.79 \\
    Qwen2.5 (5-shot + kNN + data-specific prompt)
    & 9.26 & 8.89 & 9.07
    & 6.10 & 4.76 & 5.35
    & 18.94 & 11.76 & 14.51
    & 9.47 & 10.58 & 10.00
    & 18.44 & 28.54 & 22.41
    & 34.95 & 38.72 & 36.74
    & 16.19 & 17.21 & 16.35 \\
    SRICL-Qwen2.5-14B (Our method)
    & \textbf{53.83} & \textbf{55.36} & \textbf{54.59}
    & 39.19 & 27.62 & 32.40
    & 26.69 & 20.54 & 23.21
    & 35.38 & 37.40 & 36.36
    & \textbf{61.30} & \textbf{61.07} & \textbf{61.18}
    & 44.08 & 38.83 & 41.29
    & \textbf{43.41} & \textbf{40.14} & \textbf{41.51} \\
    \midrule
    \multicolumn{22}{l}{\textbf{RELAX / NER-STYLE}} \\
    zero-shot+specific (GPT-3.5)
    & 21.43 & 9.68 & 13.33
    & 5.92 & 13.80 & 8.29
    & 1.54 & 4.76 & 2.326
    & 7.89 & 2.10 & 3.315
    & 3.08 & 11.94 & 4.90
    & 5.11 & 3.42 & 4.10
    & 6.42 & 7.62 & 6.04 \\
    5-shot
    & 74.47 & 61.95 & 67.63
    & 11.59 & 38.09 & 17.78
    & 25.9 & 60.14 & 
    & 70.82 & 67.74 & 69.24
    & 47.68 & 57.94 & 52.31
    & 26.64 & 63 & 37.44
    & 70.03 & 58.14 & 57.01 \\
    +kNN
    & 71.67 & 61.58 & 66.25
    & 23.53 & 22.22 & 22.86
    & 92.52 & 69.72 & 79.52
    & 97.06 & 75 & 84.62
    & 42.10 & 50.71 & 46.01
    & 68.76 & 60.27 & 64.23
    & 56.52 & 56.58 & 60.58 \\
    +kNN+specific
    & 26.77 & 60.18 & 37.06
    & 30.15 & 59.41 & 40.00
    & 69.82 & 63.50 & 66.51
    & 55.31 & 48.48 & 51.67
    & 44.91 & 55.09 & 49.48
    & 68.53 & 69.90 & 69.21
    & 42.21 & 59.43 & 52.32 \\
    Qwen2.5 (5-shot + kNN + data-specific prompt) 
    & 33.14 & 31.83 & 32.48
    & 45.12 & 35.24 & 39.57
    & 67.15 & 43.48 & 52.78
    & 88.07 & 98.38 & 93.01
    & 36.57 & 56.84 & 44.51
    & 50.96 & 56.71 & 53.68
    & 53.50 & 53.75 & 52.67 \\
    Llama-3-8B (SRICL- Our Method)
    & 9.26 & 8.89 & 9.07
    & 6.10 & 4.76 & 5.35
    & 18.94 & 11.76 & 14.51
    & 9.47 & 10.58 & 10.00
    & 18.44 & 28.54 & 22.41
    & 34.95 & 38.72 & 36.74
    & 16.19 & 17.21 & 16.35 \\
    SRICL-Qwen2.5-14B (Our method)
    & \textbf{75.22} & \textbf{77.43} & \textbf{76.31}
    & \textbf{63.51} & 44.76 & \textbf{52.51}
    & 63.84 & \textbf{65.53} & 64.67
    & 73.13 & \textbf{99.63} & \textbf{85.63}
    & \textbf{78.66} & \textbf{81.72} & \textbf{80.16}
    & 24.67 & 40.14 & 30.56
    & \textbf{63.17} & \textbf{68.53} & \textbf{64.97} \\
    \bottomrule
  \end{tabular}%
  }
\end{table*}

\noindent\textbf{Across-dataset performance.}
On \textit{SkillSpan}, our system achieves \textbf{54.59} F1 (P=53.83, R=55.36), substantially higher than GPT-3.5 few-shot and the LLaMA-3-8B variant.
On \textit{Sayfullina}, we obtain \textbf{61.18} F1, indicating strong robustness on fully skill-dense sentences.
For \textit{Kompetencer} (Danish) and \textit{Green} (long-tail ecological domain), our method delivers competitive F1 (32.40 and 23.21, respectively), outperforming GPT-3.5 variants while remaining below JobBERT on certain sets, reflecting the domain/language shift difficulty.
Averaged over all datasets, our system reaches \textbf{41.51} F1, surpassing all GPT-3.5 configurations in Table~\ref{tab:main_experiment}.

\noindent\textbf{Open-source vs.\ closed-source baselines.}
Compared to GPT-3.5 zero-/few-shot, the gains are consistent across languages, especially on datasets with longer sentences or heavier use of domain terminology (e.g., \textit{SkillSpan}, \textit{GNEHM}).

\begin{table}[t]
  \centering
  \caption{Ablation on SkillSpan and Sayfullina with Qwen2.5-14B.
  Scores are skill-level precision (P), recall (R), and F1. Best in each dataset is in \textbf{bold}. All are tested 0-shot.}
  \label{tab:ablation-two-datasets}
  \resizebox{\linewidth}{!}{
    \begin{tabular}{@{}lcccccc@{}}
    \toprule
     & \multicolumn{3}{c}{\textbf{SKILLSPAN}} & \multicolumn{3}{c}{\textbf{SAYFULLINA}} \\ \cmidrule(l){2-7} 
    \textbf{Method (Qwen2.5-14B)} & P & R & F1 & P & R & F1 \\ \midrule
    B8 (SFT+RAG, specific) & 0.508 & \textbf{0.571} & 0.538 & 0.480 & 0.400 & \textbf{0.320} \\
    w/o SFT (Base+RAG, specific) & 0.123 & 0.338 & 0.180 & \textbf{0.670} & 0.040 & 0.040 \\
    w/o RAG (SFT, specific) & \textbf{0.524} & 0.568 & \textbf{0.545} & 0.450 & \textbf{0.430} & \textbf{0.320} \\
    w/o specific (SFT+RAG, general) & 0.478 & 0.564 & 0.518 & 0.380 & \textbf{0.430} & 0.280 \\ \bottomrule
    \end{tabular}
}
\end{table}

\subsection{Ablation Study}
\begin{figure}[t]
  \centering
  \includegraphics[width=\columnwidth]{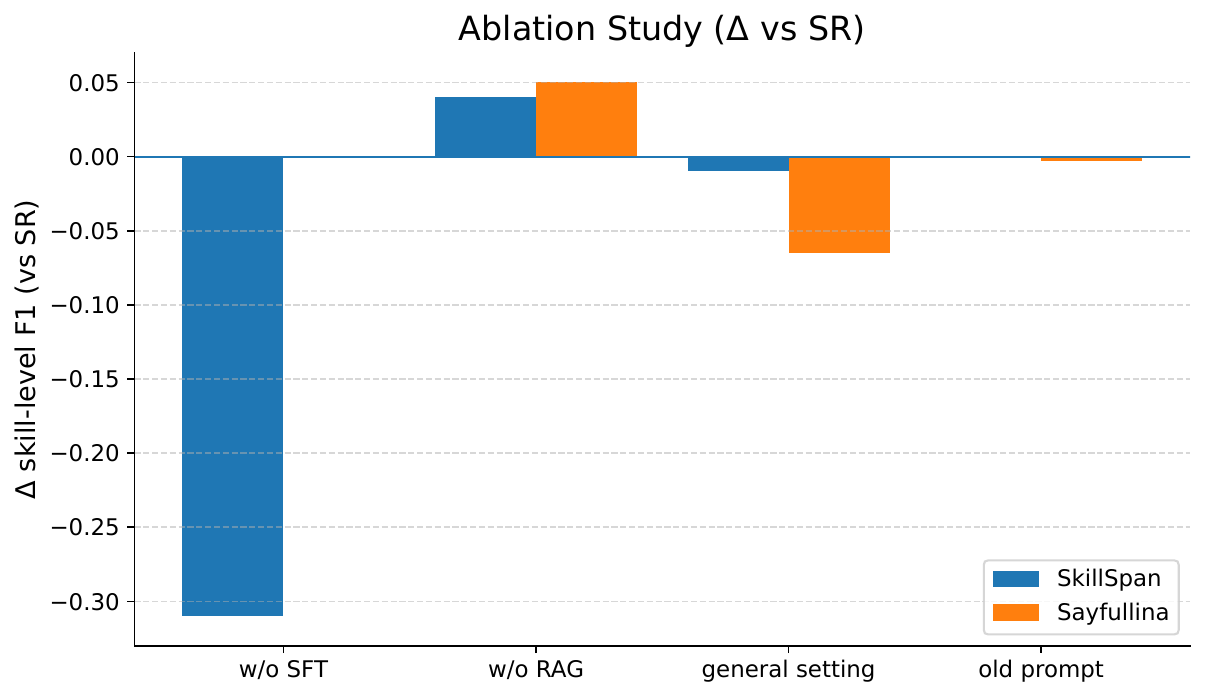}
  \caption{Ablation}
  \label{fig:ablation-delta}
\end{figure}

We ablate four components on Qwen2.5-14B (Fig.~\ref{fig:ablation-delta}, Table~\ref{tab:ablation-two-datasets}):
\emph{(i) w/o SFT}, 
\emph{(ii) w/o RAG}, 
\emph{(iii) w/o data-specific prompt} (generic prompt), and 
\emph{(iv) old prompt}.
We report the absolute change in skill-level F1 ($\Delta$) w.r.t.\ \framework{} (our full model).

\textbf{SFT.} Removing SFT yields the largest drop on \textsc{SkillSpan} ($-0.31$), while the change on \textsc{Sayfullina} is negligible ($\approx 0$). This suggests SFT primarily stabilizes span boundaries on datasets with longer, multiword skills (cf. precision gains in Fig.~\ref{fig:ablation-delta}).

\textbf{RAG.} Disabling retrieval slightly \emph{increases} skill-level F1 on both datasets ($+0.04$–$+0.05$). However, Fig.~\ref{fig:ablation-delta} shows a precision–recall trade-off: with RAG, \textsc{SkillSpan} gains recall but loses precision, yielding an approximately unchanged token-level F1. On \textsc{Sayfullina}, token-level F1 drops markedly \emph{without} RAG, indicating retrieval improves boundary legality on noisier sentences, even if span-level aggregation can mask these gains.

\textbf{Prompting.} Replacing data-specific prompts with a generic template reduces F1 ($-0.06$ on \textsc{Sayfullina}, $-0.01$ on \textsc{SkillSpan}), reflecting boundary drift on technical phrases. Using an older template is roughly neutral on \textsc{SkillSpan} ($+0.003$) but negative on \textsc{Sayfullina} ($-0.045$).

\subsubsection*{Takeaways}
(i) \textit{SFT} is the primary driver of precision and boundary stability on span-heavy corpora; 
(ii) \textit{data-specific prompting} provides consistent gains; 
(iii) \textit{RAG} is dataset-dependent—typically improving token-level robustness/recall while occasionally lowering span-level F1 when precision costs outweigh recall gains.

\subsection{Sensitivity to Retrieval}
We study RAG@k (k=\{0,3,10\}) on a held-out subset (Figure.3).
Increasing $k$ improves recall up to a saturation point, after which latency grows faster than the marginal F1 gain.
This suggests a small $k$ with quality-filtered sources (train-style examples + ESCO definitions) is a good operating point for production.

\begin{figure}[t]
  \centering
  \includegraphics[width=\columnwidth]{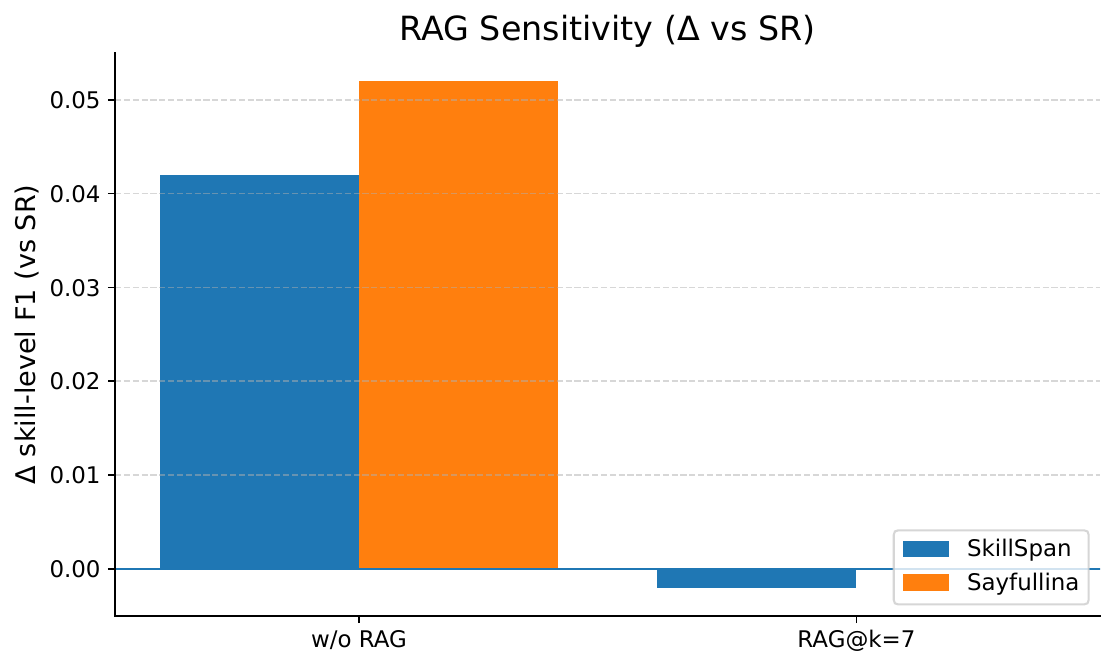}
  \caption{Rag Sensitivity}
  \label{fig:block}
\end{figure}

%% file: 5RelatedWork.tex
\section{Related Work}
\label{sec:prior}
Skill extraction for recruitment analytics is typically cast as span-level NER with BIO tags on job ads, while generative IE with LLMs is rapidly advancing \cite{xu2024large}. Methodologically, prior work clusters into three lines: \emph{supervised neural NER} \cite{conneau-etal-2020-xlm-r}, \emph{prompt-based LLMs} (e.g., CoT, self-consistency) \cite{wei2022cot,wang2022self}, and \emph{retrieval-augmented} generation \cite{lewis2020rag}. For skill extraction specifically, studies examine in-context learning \cite{nguyen-etal-2024-rethinking} and instruction-driven, text-to-structure IE frameworks \cite{lu-etal-2022-unified,wang2023instructuie}. 

Despite progress, span-level deployment remains hard: supervised models are label-hungry and brittle across domains; prompting pipelines risk hallucinations and unstable, token-aligned boundaries; retrieval alone offers limited BIO-level control. Post-hoc verification and constrained decoding improve structural fidelity \cite{kim2024verifiner,geng2024sketch,he2023unleashing}, yet are seldom paired with \emph{multi-source, skill-aware grounding} tailored to job ads. In practice, skill taxonomies such as ESCO are still under-used for span disambiguation and boundary regularization in end-to-end generation \cite{ESCO_Main_Release,ESCO_Green_Skills}.


%% file: 6Conclusion.tex
\section{CONCLUSION}
\label{sec:majhead}

(1) A modular, verification-aware framework for job-skill extraction that couples compact SFT with \textbf{multi-source RAG} and task-specific prompting. 
(2) A thorough evaluation on six public datasets with open/closed LLM baselines and ablations quantifying the marginal utility of classifier/CoT/RAG components. 
(3) Practical guidance for \emph{low-resource, multi-domain} deployments: an open-model recipe competitive with supervised sequence labelers while remaining controllable and auditable via retrieved evidence.

%% file: 7Appendix.tex
\section{Supplementary Ablation Evidence}

\begin{figure}[t]
    \centering
    \includegraphics[width=\columnwidth]{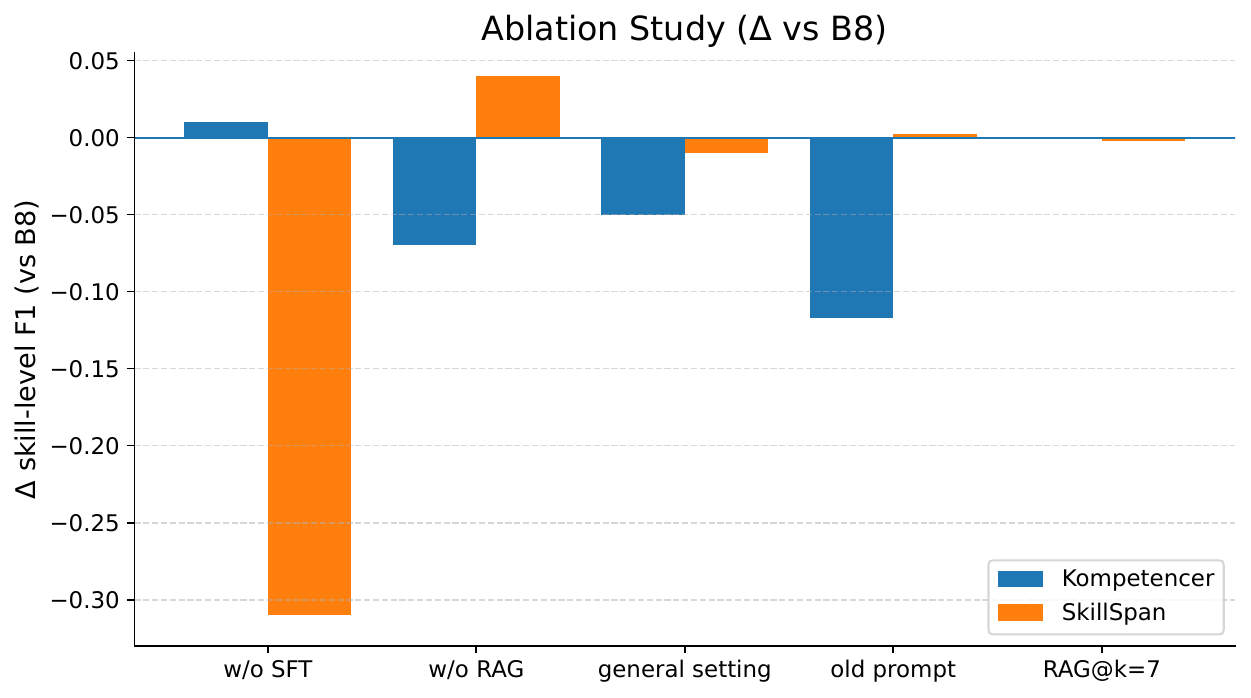}
    \caption{Relative change in skill-level F1 with respect to the B8 setting on SkillSpan and Kompetencer.}
    \label{fig:appendix_ablation_both}
\end{figure}

To complement Table~\ref{tab:ablation-two-datasets}, Fig.~\ref{fig:appendix_ablation_both} shows the relative change in skill-level F1 with respect to the B8 configuration on two benchmarks. The figure highlights that the effect of individual components is dataset-dependent: removing SFT leads to the largest drop overall, while the influence of retrieval and prompt variants varies across datasets. This supplementary view is consistent with our main conclusion that SFT is the primary driver of boundary stability, whereas RAG and prompt design mainly affect robustness and recall in a dataset-specific manner.

\section{Supplementary Multi-Metric Visualization}

\begin{figure}[t]
    \centering
    \includegraphics[width=0.92\columnwidth]{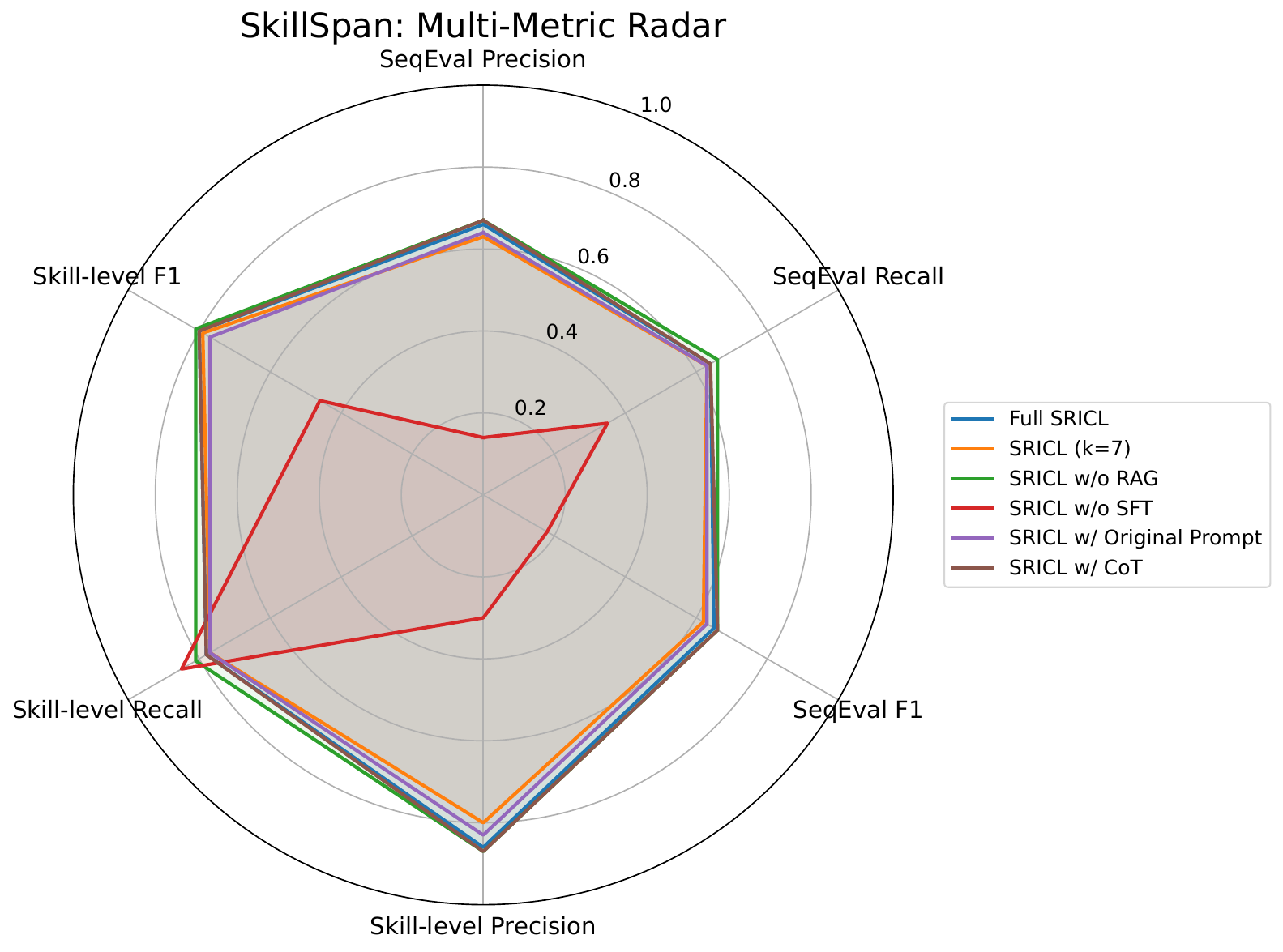}
    \caption{Radar-chart comparison of multi-metric performance on SkillSpan across SRICL variants. The figure summarizes sequence-level and skill-level precision, recall, and F1, providing an overall view of the trade-offs among different ablation settings.}
    \label{fig:skillspan_radar}
\end{figure}

To complement the ablation tables and bar-chart comparisons in the main text, Fig.~\ref{fig:skillspan_radar} presents a radar-chart view of multi-metric performance on SkillSpan. Specifically, we compare six configurations, including the full SRICL setting, the k=7 retrieval variant, and four ablated variants without RAG, without SFT, with the original prompt, and with CoT. The figure summarizes performance across both sequence-level and skill-level metrics, including SeqEval precision, SeqEval recall, SeqEval F1, Skill-level precision, Skill-level recall, and Skill-level F1.

Overall, the full SRICL configuration exhibits a relatively balanced and competitive performance profile across all six metrics. Removing SFT leads to the most substantial degradation, especially on sequence-level precision and F1, confirming that SFT is the primary driver of boundary stability and structured extraction quality. By contrast, removing RAG or altering the prompting strategy produces more moderate and metric-dependent changes. This supplementary visualization is consistent with the main findings reported in the ablation section and provides an intuitive overview of the trade-offs among different configurations.

%% file: 9Reference.bib
@inproceedings{zhang2022skillspan,
  title     = {SkillSpan: Hard and Soft Skill Extraction from English Job Postings},
  author    = {Zhang, Mike and Jensen, Kristian and Sonniks, Sif and Plank, Barbara},
  booktitle = {Proceedings of the 2022 Conference of the North American Chapter of the Association for Computational Linguistics: Human Language Technologies},
  year      = {2022},
  pages     = {4962--4984},
  address   = {Seattle, United States},
  publisher = {Association for Computational Linguistics},
  doi       = {10.18653/v1/2022.naacl-main.366}
}

@article{lewis2020rag,
  title   = {Retrieval-Augmented Generation for Knowledge-Intensive NLP Tasks},
  author  = {Lewis, Patrick and Perez, Ethan and Piktus, Aleksandra and Petroni, Fabio and Karpukhin, Vladimir and Goyal, Naman and K{\"u}ttler, Heinrich and Lewis, Mike and Yih, Wen-tau and Rockt{\"a}schel, Tim and Riedel, Sebastian and Kiela, Douwe},
  journal = {Advances in Neural Information Processing Systems},
  volume  = {33},
  pages   = {9459--9474},
  year    = {2020}
}

@article{wei2022cot,
  title={Chain-of-Thought Prompting Elicits Reasoning in Large Language Models},
  author={Wei, Jason and Wang, Xuezhi and Schuurmans, Dale and others},
  journal={arXiv:2201.11903},
  year={2022}
}

@inproceedings{conneau-etal-2020-xlm-r,
  title={{U}nsupervised Cross-lingual Representation Learning at Scale},
  author={Conneau, Alexis and Khandelwal, Kartikay and Goyal, Naman and Chaudhary, Vishrav and Wenzek, Guillaume and Guzm{\'a}n, Fransisco and others},
  booktitle={ACL},
  year={2020}
}

@inproceedings{manakul2023selfcheckgpt,
  title     = {SelfCheckGPT: Zero-Resource Black-Box Hallucination Detection for Generative Large Language Models},
  author    = {Manakul, Potsawee and Liusie, Adian and Gales, Mark},
  booktitle = {Proceedings of the 2023 Conference on Empirical Methods in Natural Language Processing},
  year      = {2023},
  pages     = {9004--9017},
  address   = {Singapore},
  publisher = {Association for Computational Linguistics},
  doi       = {10.18653/v1/2023.emnlp-main.557}
}

@article{xu2024large,
  title={Large language models for generative information extraction: A survey},
  author={Xu, Derong and Chen, Wei and Peng, Wenjun and Zhang, Chao and Xu, Tong and Zhao, Xiangyu and Wu, Xian and Zheng, Yefeng and Wang, Yang and Chen, Enhong},
  journal={Frontiers of Computer Science},
  volume={18},
  number={6},
  pages={186357},
  year={2024},
  publisher={Springer}
}

@inproceedings{nguyen-etal-2024-rethinking,
    title = "Rethinking Skill Extraction in the Job Market Domain using Large Language Models",
    author = "Nguyen, Khanh  and
      Zhang, Mike  and
      Montariol, Syrielle  and
      Bosselut, Antoine",
    editor = "Hruschka, Estevam  and
      Lake, Thom  and
      Otani, Naoki  and
      Mitchell, Tom",
    booktitle = "Proceedings of the First Workshop on Natural Language Processing for Human Resources (NLP4HR 2024)",
    month = mar,
    year = "2024",
    address = "St. Julian{'}s, Malta",
    publisher = "Association for Computational Linguistics",
    url = "https://aclanthology.org/2024.nlp4hr-1.3/",
    pages = "27--42"
}

@inproceedings{senger2024deep,
  title     = {Deep Learning-based Computational Job Market Analysis: A Survey on Skill Extraction and Classification from Job Postings},
  author    = {Senger, Elena and Zhang, Mike and van der Goot, Rob and Plank, Barbara},
  booktitle = {Proceedings of the First Workshop on Natural Language Processing for Human Resources (NLP4HR 2024)},
  year      = {2024},
  pages     = {1--15},
  address   = {St. Julian's, Malta},
  publisher = {Association for Computational Linguistics}
}

@inproceedings{lu-etal-2022-unified,
    title = "Unified Structure Generation for Universal Information Extraction",
    author = "Lu, Yaojie  and
      Liu, Qing  and
      Dai, Dai  and
      Xiao, Xinyan  and
      Lin, Hongyu  and
      Han, Xianpei  and
      Sun, Le  and
      Wu, Hua",
    editor = "Muresan, Smaranda  and
      Nakov, Preslav  and
      Villavicencio, Aline",
    booktitle = "Proceedings of the 60th Annual Meeting of the Association for Computational Linguistics (Volume 1: Long Papers)",
    month = may,
    year = "2022",
    pages = "5755--5772",
}

@article{wang2023instructuie,
  title={Instructuie: Multi-task instruction tuning for unified information extraction},
  author={Wang, Xiao and Zhou, Weikang and Zu, Can and Xia, Han and Chen, Tianze and Zhang, Yuansen and Zheng, Rui and Ye, Junjie and Zhang, Qi and Gui, Tao and others},
  journal={arXiv preprint arXiv:2304.08085},
  year={2023}
}

@article{shinn2023reflexion,
  title={Reflexion: Language agents with verbal reinforcement learning, 2023},
  author={Shinn, Noah and Cassano, Federico and Labash, Beck and Gopinath, Ashwin and Narasimhan, Karthik and Yao, Shunyu},
  journal={URL https://arxiv. org/abs/2303.11366},
  volume={1},
  year={2023}
}

@article{madaan2023self,
  title={Self-refine: Iterative refinement with self-feedback},
  author={Madaan, Aman and Tandon, Niket and Gupta, Prakhar and Hallinan, Skyler and Gao, Luyu and Wiegreffe, Sarah and Alon, Uri and Dziri, Nouha and Prabhumoye, Shrimai and Yang, Yiming and others},
  journal={Advances in Neural Information Processing Systems},
  volume={36},
  pages={46534--46594},
  year={2023}
}

@inproceedings{wang2022self,
  title     = {Self-Consistency Improves Chain of Thought Reasoning in Language Models},
  author    = {Wang, Xuezhi and Wei, Jason and Schuurmans, Dale and Le, Quoc and Chi, Ed and Narang, Sharan and Chowdhery, Aakanksha and Zhou, Denny},
  booktitle = {The Eleventh International Conference on Learning Representations},
  year      = {2023}
}

@inproceedings{kim2024verifiner,
  title     = {VerifiNER: Verification-augmented NER via Knowledge-grounded Reasoning with Large Language Models},
  author    = {Kim, Seoyeon and Seo, Kwangwook and Chae, Hyungjoo and Yeo, Jinyoung and Lee, Dongha},
  booktitle = {Proceedings of the 62nd Annual Meeting of the Association for Computational Linguistics (Volume 1: Long Papers)},
  year      = {2024},
  pages     = {2441--2461},
  address   = {Bangkok, Thailand},
  publisher = {Association for Computational Linguistics}
}

@article{he2023unleashing,
  title={Unleashing the true potential of sequence-to-sequence models for sequence tagging and structure parsing},
  author={He, Han and Choi, Jinho D},
  journal={Transactions of the Association for Computational Linguistics},
  volume={11},
  pages={582--599},
  year={2023},
  publisher={MIT Press One Broadway, 12th Floor, Cambridge, Massachusetts 02142, USA~…}
}

@article{geng2024sketch,
  title={Sketch-guided constrained decoding for boosting blackbox large language models without logit access},
  author={Geng, Saibo and D{\"o}ner, Berkay and Wendler, Chris and Josifoski, Martin and West, Robert},
  journal={arXiv preprint arXiv:2401.09967},
  year={2024}
}

@article{ma2023large,
  title={Large language model is not a good few-shot information extractor, but a good reranker for hard samples!},
  author={Ma, Yubo and Cao, Yixin and Hong, YongChing and Sun, Aixin},
  journal={arXiv preprint arXiv:2303.08559},
  year={2023}
}

@misc{ESCO_Main_Release,
  author       = {{European Commission, Directorate-General for Employment, Social Affairs and Inclusion}},
  title        = {{ESCO: European Skills, Competences, Qualifications and Occupations} (Main classification, v1.2)},
  howpublished = {Online database},
  year         = {2024},
  note         = {Available: https://ec.europa.eu/esco/ \; Accessed: 2025-09-17}
}

@misc{ESCO_Green_Skills,
  author       = {{European Commission}},
  title        = {{ESCO Escopedia: Green skills and related knowledge concepts}},
  howpublished = {Web page},
  year         = {2024},
  note         = {Available: https://ec.europa.eu/esco/ \; Accessed: 2025-09-17}
}
